\documentclass[11pt]{article}

\usepackage[margin=1in]{geometry}
\usepackage{threeparttable}
\usepackage{multirow}
\usepackage{pgfplots}
\usepackage{subfigure}
\usepackage{booktabs}
\usepackage{authblk}
\usepackage{hyperref}
\usepackage{rotating}
\usepackage{amsmath}
\usepackage{amssymb} % defines \mathbb

\title{Structure-Aware Epistemic Uncertainty Quantification for Neural Operator PDE Surrogates}

\usepackage{authblk} % (in preamble)
\usepackage{hyperref} % (in preamble, optional)

\newcommand{\corr}{\thanks{Corresponding authors.}} % no emails

\author[1]{Haoze Song}
\author[1]{Zhihao Li}
\author[1]{Mengyi Deng}
\author[1]{Xin Li}
\author[1]{Duyi Pan}
\author[1,2]{Zhilu Lai\corr}
\author[1,2]{Wei Wang\textsuperscript{*}} % same mark as Zhilu Lai

\affil[1]{The Hong Kong University of Science and Technology (Guangzhou), Guangzhou, China}
\affil[2]{The Hong Kong University of Science and Technology, Hong Kong SAR, China}

\date{} % empty date

\begin{document}
\maketitle

\begin{abstract}
Neural operators (NOs) provide fast, resolution-invariant surrogates for mapping input fields to PDE solution fields, but their predictions can exhibit significant epistemic uncertainty due to finite data, imperfect optimization, and distribution shift. For practical deployment in scientific computing, uncertainty quantification (UQ) must be both computationally efficient and spatially faithful, i.e., uncertainty bands should align with the localized residual structures that matter for downstream risk management. We propose a structure-aware epistemic UQ scheme that exploits the modular anatomy common to modern NOs (lifting–propagation–recovering). Instead of applying unstructured weight perturbations (e.g., naive dropout) across the entire network, we restrict Monte Carlo sampling to a module-aligned subspace by injecting stochasticity only into the lifting module, and treat the learned solver dynamics (propagation and recovery) as deterministic. We instantiate this principle with two lightweight lifting-level perturbations, including channel-wise multiplicative feature dropout and a Gaussian feature perturbation with matched variance, followed by standard calibration to construct uncertainty bands. Experiments on challenging PDE benchmarks (including discontinuous-coefficient Darcy flow and geometry-shifted 3D car CFD surrogates) demonstrate that the proposed structure-aware design yields more reliable coverage, tighter bands, and improved residual–uncertainty alignment compared with common baselines, while remaining practical in runtime.
\end{abstract}

% Keywords (plain article doesn't have \keywords)

\section{Introduction}

Uncertainty quantification (UQ) in supervised learning aims to characterize the distribution of model outputs conditioned on given inputs. In surrogate modeling of partial differential equations (PDEs), UQ can be as important as point prediction: well-calibrated uncertainty bands can indicate where and by how much a surrogate may deviate from a high-fidelity numerical solver, providing a practical notion of risk and enabling more reliable deployment in accelerated pipelines such as aerospace CFD design \cite{24:PINO_Aerospace, 26:GINO_Aerospace}, electromagnetic system simulation \cite{22:NO_electromagnetic, 23:GNOT}, and safety-critical monitoring in the nuclear industry \cite{23:FNO_nuclear}.

As fast, resolution-invariant surrogates that learn mappings between function spaces from data, neural operators (NOs) \cite{19:deeponet, 21:fno, 24:transolver} are widely adopted because they can be deployed efficiently and reused across discretizations. However, with finite training data, limited optimization budgets, and potential distribution shift at inference time, NO predictions are inherently uncertain. In many PDE settings, the underlying simulator is effectively deterministic (i.e., aleatoric noise is negligible), so the dominant uncertainty in practical NO deployment is epistemic: uncertainty about the learned operator induced by limited data and imperfect training.

\subsection{Epistemic Uncertainty in NOs}

An NO $\mathcal{G}[\cdot; \theta]:\mathbf{A}\mapsto \hat{\mathbf{U}}$ approximates a target operator $\mathcal{G}$ that maps discretized input fields $\mathbf{A}\in\mathbb{R}^{N\times d_a}$ to discretized solution fields $\mathbf{U}\in\mathbb{R}^{N\times d_u}$, using a finite dataset $\mathcal{D}=\{(\mathbf{A}_m,\mathbf{U}_m)\}_{i=1}^M$. In a Bayesian operator-learning view, the weights $w$ are treated as random after observing $\mathcal{D}$, inducing a posterior $p(w\mid\mathcal{D})$ and a predictive distribution $p(\hat{{U}}\mid \mathbf{A},\mathcal{D})$. Posterior predictive statistics derived from this distribution are then used to construct uncertainty bands \cite{16:dropout, 17:DeepEnsemble, 24:uncertainty_workshop_NO}.

When the underlying PDE solver is effectively deterministic, epistemic uncertainty is primarily manifested as the variability of $\mathcal{G}[\mathbf{A}; \theta]$ under plausible weights $\theta\sim p(\theta\mid\mathcal{D})$. In practice, this uncertainty is shaped by several factors, including (i) optimization heuristics that bias training toward a particular local optimum, (ii) finite compute budgets and regularization choices that restrict the set of reachable solutions, and (iii) out-of-distribution (OOD) inputs that probe under-constrained regions of the learned operator.

\subsection{Previous Quantification Designs} \label{Section_1_2}

Estimating the high-dimensional posterior $p(\theta\mid\mathcal{D})$ is intractable for NOs \cite{20:Bayesian_Deep_Learning}, so existing approaches \cite{16:dropout, 17:DeepEnsemble, 24:uncertainty_workshop_NO, 24:CP_NO} rely on practical sampling and approximation schemes to estimate posterior predictive quantities, yielding predictive statistical fields such as variances, quantiles, and credible intervals. Recent approaches include:

\paragraph{Deep Ensembles.}
Deep Ensembles \cite{17:DeepEnsemble} train multiple models from different random initializations and statistically aggregate their predictions, treating the resulting variability as samples from an implicit distribution over plausible solution weights. They are often robust and empirically strong, but incur substantial training cost and still do not exploit which parts of the operator are responsible for uncertainty.

\paragraph{Laplace Approximation.} Laplace methods \cite{21:Laplace_UAI, 25:Laplace_NO} approximate the posterior $p(\theta\mid\mathcal{D})$ locally around a trained solution $\hat{\theta}$ with a Gaussian whose covariance is given by (an approximation to) the inverse Hessian of the negative log posterior evaluated at $\hat{\theta}$. For overparameterized neural operators, full-curvature Laplace is typically computationally prohibitive and can be numerically fragile \cite{18:laplaceApp_NN}. A common simplification is {last-layer Laplace} \cite{20:Last_L_Laplace}, which is cheaper and more stable, but primarily captures uncertainty in the final linear readout (i.e., how to combine learned features) while neglecting uncertainty in the learned representations. However, since NO residuals can vary substantially across the output domain, epistemic error often stems from imperfect residual representations rather than merely mis-weighted features. This feature-level mismatch induces a calibration tension: achieving higher residual coverage often requires overly inflated (wide) intervals, whereas sharper intervals tend to suffer from under-coverage.

\paragraph{MCDropout.} MCDropout \cite{16:dropout} injects randomness by sampling dropout masks at inference time and uses Monte Carlo forward passes to estimate posterior predictive moments. It is simple, scalable, and easy to integrate into existing neural operators. However, the induced perturbations in weight space are dictated by the dropout mechanism and therefore do not distinguish between parameter dimensions, effectively applying a uniform, unstructured stochastic perturbation across the network. When inserted naively, dropout can degrade predictive accuracy and lead to overly conservative, wide uncertainty bands: random masking may suppress channels or neurons that are critical for signal propagation in neural operators, producing samples in which essential intermediate features are partially removed. This can bias the Monte Carlo predictive mean and inflate predictive variance due to increased variability across sampled predictions \cite{25:EpistemicIncomplete}.

\begin{table}[ht]
\centering
\caption{Comparison of common epistemic UQ methods in neural operators.}
\label{tab:uq_methods_compare}
\setlength{\tabcolsep}{8pt}
\renewcommand{\arraystretch}{1.15}

\resizebox{0.7\textwidth}{!}{%
\begin{tabular}{lcc}
\toprule
\textbf{Method} & \textbf{Training Runs} & \textbf{Inference Runs} \\
\midrule
Deep Ensembles \cite{17:DeepEnsemble}        & $10$ & $1$ \\
Laplace Approximation \cite{25:Laplace_NO} & $1$  & -- \\
MC Dropout \cite{16:dropout}            & $1$  & $10$--$1000$ \\
\bottomrule
\end{tabular}%
}
\end{table}

We summarize the computational cost of these epistemic UQ methods in Table~\ref{tab:uq_methods_compare} for deploying a neural operator (NO) with high-coverage uncertainty bands. Since NO inference is typically far cheaper than training, the overhead of tens to hundreds of stochastic forward passes for MCDropout is often modest relative to the additional training required by Deep Ensembles. For Laplace approximation, the dominant overhead usually arises from forming and solving with an (approximate) Hessian at the trained solution, which can be both computationally expensive for large NOs when applied beyond highly structured approximations.

\subsection{Our Structure-aware Design}

While previous structure-agnostic posterior approximations of $p(\theta\mid\mathcal{D})$ often spend most of their sampling budget exploring parameter dimensions that contribute little or even harm the alignment between uncertainty bands and residual structures, we propose a structure-aware UQ scheme for neural operators. Concretely, we perform Monte Carlo integration by sampling only a module-aligned subset of parameters, $\theta_{\mathcal S}\sim p(\theta_{\mathcal{S}}\mid\mathcal D)$, where $w_{\mathcal{S}}$ corresponds to a designated operator module $\mathcal{S}$. This design is motivated by a key property of highly structured neural operators: parameters within the same module typically play similar functional roles and affect the final prediction through shared mechanisms. Ignoring such module semantics can make posterior sampling both inefficient and misleading. In particular, unstructured perturbations of sensitive parameters may produce samples with biased predictive means and large residuals, leading to overly conservative, inflated uncertainty bands; meanwhile, perturbing parameters whose induced features are (locally/globally) orthogonal to the residual field wastes sampling effort and provides little benefit for aligning uncertainty bands with residuals.

\paragraph{Contributions.} Leveraging the distinct roles of the three core components in neural operators, {Lifting}, {Propagation}, and {Recovering}, we propose a UQ scheme that prioritizes structure-aligned subspaces, effectively attributing epistemic uncertainty throughout the end-to-end mapping (from input conditions to solution fields) to uncertainty in the generated lifted feature tensors. Empirically, this structure-aware design estimates residual structures between predicted and ground-truth fields with higher fidelity than existing UQ baselines. Our main contributions in this paper are summarized as follows:
\begin{itemize}
    \item \textbf{Structure-aware epistemic UQ via lifting-subspace sampling.}
    We propose a module-aligned Monte Carlo UQ strategy for neural operators that leverages the lifting--propagation--recovering decomposition. Instead of injecting stochasticity into the full parameter space, we {restrict sampling to the lifting subspace} and keep the learned solver dynamics (propagation and recovering) deterministic. This design is efficient and structure-preserving: it models epistemic uncertainty as uncertainty over the {lifted feature field} that is then propagated by a fixed remainder operator, offering an operational interpretation akin to ``feature-space initial-condition'' uncertainty.

    \item \textbf{Two plug-and-play sampling mechanisms with matched statistics.}
    We instantiate this principle with two lightweight perturbations on the lifted features: (i) channel-wise multiplicative (dropout-style) feature noise, and (ii) Gaussian feature perturbation with variance matched to the inverted-dropout scale. Both are zero-mean around the deterministic embedding, require no retraining, and add only inference-time overhead via $T$ stochastic forward passes.

    \item \textbf{Empirical evidence on challenging PDE benchmarks and OOD generalization.}
    We evaluate on discontinuous 2D Darcy flow and geometry-shifted 3D ShapeNet Car CFD surrogates, spanning representative NO backbones (e.g., spectral-propagation and attention-based variants). Across settings, the proposed method improves the coverage--bandwidth trade-off and remains competitive in runtime relative to widely used epistemic UQ baselines, including MC Dropout, Laplace-style approximations, deep ensembles, and input perturbation. Comprehensive comparisons and dropout-focused ablations further indicate that our structure-aware sampling yields {stable} uncertainty maps that better track residual fields attributable to epistemic uncertainty.
\end{itemize}

\section{Problem Setup}

\subsection{Operator Learning}\label{Section_2_1}

A PDE solution operator $\mathcal{G}$ maps an input function (e.g., initial conditions, forcing terms, or boundary conditions) $a(x)\in\mathcal{A}$ to a solution function $u(x)\in\mathcal{U}$, where $\mathcal{A}=\mathcal{A}(D;\mathbb{R}^{d_a})$ and $\mathcal{U}=\mathcal{U}(D;\mathbb{R}^{d_u})$ are (suitable subsets of) Banach spaces over a spatial domain $D$. Here we consider vector-valued fields $a: D\to\mathbb{R}^{d_a}$ and $u: D\to\mathbb{R}^{d_u}$ with $x\in D$.

Neural operators parameterize $\mathcal{G}$ with weights $\theta$, yielding a learned operator $\mathcal{G}_{\theta}$. In practice, we work with discretized fields on a grid $\{x_n\}_{n=1}^N\subset D$. The discretized input is a tensor $\mathbf{A}\in\mathbb{R}^{N\times d_a}$ with rows $\mathbf{A}_{n,:}=a(x_n)$, and the discretized solution is $\mathbf{U}\in\mathbb{R}^{N\times d_u}$ with rows $\mathbf{U}_{n,:}=u(x_n)$.

A typical neural operator can be decomposed into three components,
\begin{equation} \label{Equation_1}
\mathcal{G}_{\theta} \;=\; \mathcal{Q}_{\theta_{\mathcal{Q}}}\circ \mathcal{M}_{\theta_{\mathcal{M}}}\circ \mathcal{P}_{\theta_{\mathcal{P}}},
\end{equation}
where: (i) the {lifting} operator $\mathcal{P}_{\theta_{\mathcal{P}}}: \mathbf{A}\mapsto \mathbf{V}_0\in\mathbb{R}^{N\times d_v}$ embeds the input into a latent feature field;
(ii) the {propagation} operator $\mathcal{M}_{\theta_{\mathcal{M}}}: \mathbf{V}_0\mapsto \mathbf{V}_L$ iteratively transforms latent features via structured linear operators and nonlinear activations, typically containing the majority of parameters and determining much of the approximation capacity \cite{21:universal_no}; and
(iii) the {recovering} operator $\mathcal{Q}_{\theta_{\mathcal{Q}}}: \mathbf{V}_L\in\mathbb{R}^{N\times d_v}\mapsto \hat{\mathbf{U}}$ maps final latent features back to the output space, commonly implemented as a linear layer.

When explicit PDE priors are unavailable, neural operators are often trained in a purely data-driven manner by minimizing losses such as mean squared error (MSE) or relative $L_2$ error (defined in Appendix~\ref{Appendix_loss}). Throughout this paper, we focus on neural operators that follow the above architectural composition and are trained under this purely data-driven paradigm; operator-learning models outside this framework (e.g., DeepONet-style architectures \cite{19:deeponet}) are not considered.

\subsection{Epistemic Uncertainty Quantification} \label{Section_2_2}

From a Bayesian perspective, epistemic uncertainty in neural operators is reflected in how well we can characterize the posterior over weights and translate it into calibrated predictive statistics. Given an input $a$, let $\hat{u}(x)$ denote the predicted solution and define the residual field
\begin{equation} \label{Equation_2}
r(x) := (\hat{u}-u)(x).
\end{equation}
We consider a nonnegative uncertainty-band field $\mathrm{band}(x)\ge 0$ that defines the pointwise prediction interval
\begin{equation}
\hat{u}(x)\pm \mathrm{band}(x).
\end{equation}

For vector-valued outputs, we apply the band channel-wise and compute coverage per channel; a location $x_n$ is counted as covered only if all channels are covered.

Prior works \cite{24:uncertainty_workshop_NO,24:CP_NO, 25:Laplace_NO} evaluate UQ methods using (i) coverage rate ($\mathrm{C.R.}$) and (ii) average bandwidth ($\mathrm{Avg.\ B.W.}$). For a test set $\{(a_m,u_m)\}_{m=1}^{M_{\mathrm{test}}}$ and grid points $\{x_n\}_{j=1}^N$, considering one channel $d_u = 1$, define the indicator $\mathbb{I}[\cdot]\in\{0,1\}$. The case-wise coverage rate for a test case $m$ is
\begin{equation} \label{Equ_4}
\mathrm{C.R.}_{m}:=\frac{1}{N}\sum_{n=1}^{N}\mathbb{I}\!\left[\ \| \hat{u}_m(x_n)-u_m(x_n)\|_2 \le \mathrm{band}_m(x_n)\ \right],
\end{equation}
and the average case-wise coverage over the test set is
\begin{equation}
\label{eq:coverage_pointwise}
\mathrm{Avg.\ C.R.}:=
\frac{1}{M_{\mathrm{test}}}\sum_{m=1}^{M_{\mathrm{test}}}\mathrm{C.R.}_m.
\end{equation}
The {overall} coverage aggregated over the entire test set and spatial grid is
\begin{equation}
\label{eq:coverage_global}
\mathrm{Total\ C.R.}:=\frac{1}{M_{\mathrm{test}}N}\sum_{m=1}^{M_{\mathrm{test}}}\sum_{n=1}^{N}
\mathbb{I}\!\left[\ \| \hat{u}_m(x_n)-u_m(x_n)\|_2 \le \mathrm{band}_m(x_n)\ \right].
\end{equation}
Note that for under uniform grids, $N$ is the same and thus $\mathrm{Avg.\ C.R.}$ and $\mathrm{Total\ C.R.}$ coincide.

(ii) The average (half) bandwidth is
\begin{equation}
\label{eq:avg_bandwidth}
\mathrm{Avg.\ B.W.}:=\frac{1}{M_{\mathrm{test}}N}\sum_{m=1}^{M_{\mathrm{test}}}\sum_{n=1}^{N}
\mathrm{band}_m(x_n).
\end{equation}

In the idealized limit, a band that exactly matches the pointwise residual magnitude attains $100\%$ coverage with the smallest possible width. Accordingly, better UQ methods should achieve higher coverage with tighter bands. However, in practice, even methods with high coverage and relatively low bandwidth can still fail to capture the {spatial structure} of the residual field (Section~\ref{Section_1_2}). Such misalignment yields unnecessarily conservative uncertainty in regions where the bands greatly exceed the true residuals, increasing downstream correction and design costs by triggering avoidable interventions in otherwise reliable regions.

\section{Method}

Epistemic uncertainty is typically modeled via stochastic inference over model parameters. A widely used tool is Monte Carlo (MC) integration, whose convergence rate scales as $\mathcal{O}(T^{-1/2})$ and is largely insensitive to the ambient dimension. Its main drawback is statistical noise: accurate estimates of posterior expectations may require a large number of samples.

\begin{figure*}[ht]
  \centering
  \includegraphics[width=\textwidth]{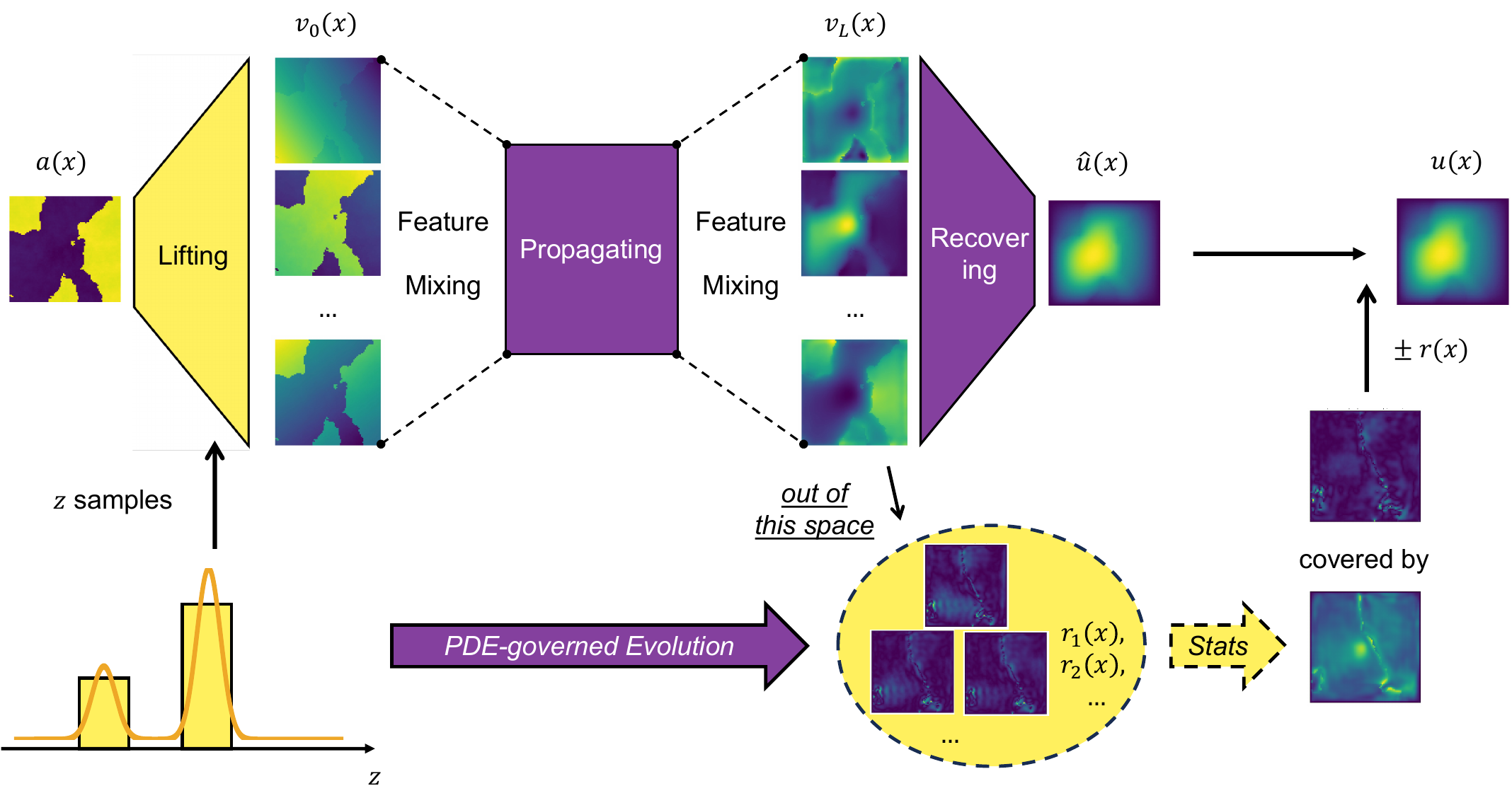}
  \caption{Overview of neural operator architectures and our structure-aware uncertainty quantification (UQ) scheme.}
  \label{fig:framework}
\end{figure*}

As illustrated in Figure~\ref{fig:framework}, our goal is to approximate posterior statistics under $p(\theta\mid\mathcal{D})$ efficiently. We exploit the modular structure common to neural operators and restrict sampling to a {module-aligned} subspace. This structure-aware restriction targets the degrees of freedom that most directly control the {epistemic residual field}, yielding more faithful and sample-efficient uncertainty estimates than perturbations of other weights.

\subsection{Structure-aware Subspace Sampling}

\begin{figure}[ht]
  \centering
  \includegraphics[width=0.8\textwidth]{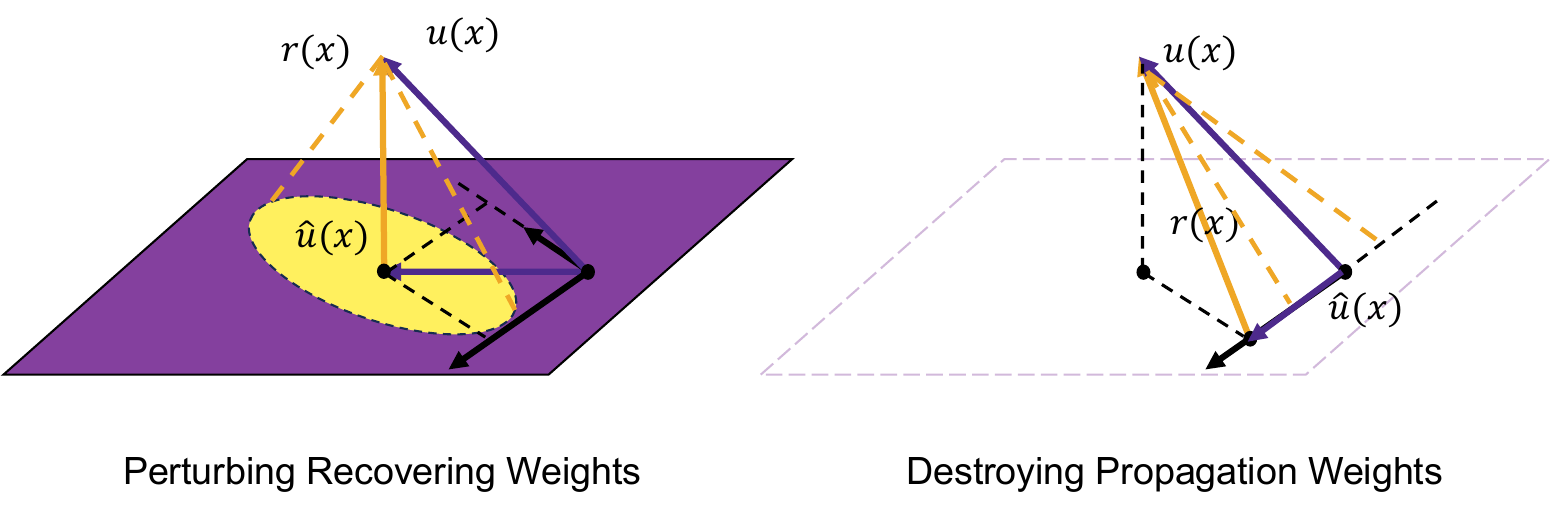}
  \caption{Sampling from other parameter groups can lead to uninformative and low-quality samples.}
  \label{fig:harm}
\end{figure}

To build a generic UQ method that respects the typical anatomy of neural operators, we group parameters into three modules (Section~\ref{Section_2_1}): lifting ($\mathcal{P}$), propagation ($\mathcal{M}$), and recovering ($\mathcal{Q}$). Crucially, not all parameter dimensions (or modules) are equally informative for epistemic UQ. Sampling an inappropriate set of parameters can lead to either (i) overly diverse residual samples that drift away from the typical residual structure (e.g., the outlier fields highlighted by the dashed region in Figure~\ref{fig:harm}), producing conservative and inflated bands, or (ii) samples that miss key residual modes, yielding miscalibration (e.g., under-coverage unless bands are widened). In short, effective epistemic UQ benefits from identifying {which parameters control which residual structures}.

\paragraph{Why sample the lifting module?}
As demonstrated in Figure~\ref{fig:harm}, sampling the recovering module $\mathcal{Q}$ (often a linear readout) primarily changes how latent feature channels are linearly combined at the output. Empirically, for a trained operator near a local optimum, the learned output feature space may be only weakly aligned with the dominant residual directions. Consequently, perturbing $\mathcal{Q}$ can introduce unnecessary variability, producing outlier fields that do not match the spatial organization of residuals.

On the other hand, the propagation module $\mathcal{M}$ typically contains most parameters and implements complex nonlocal transformations with nonlinearities. Unstructured perturbations in $\mathcal{M}$ can easily degrade predictive accuracy and induce biased or overly conservative uncertainty estimates due to uncontrolled error amplification through nonlinear layers.

These considerations motivate a structured projection of posterior sampling onto the lifting subspace: we inject stochasticity only through the lifting module and treat the remaining mapping as deterministic. Intuitively, lifting mainly prepares the initial embedding tensors (often through near-linear mixing of input channels) and does not itself represent the PDE evolution operator \cite{21:universal_no}. In contrast, $\mathcal{M}$ and $\mathcal{Q}$ encode the learned solver dynamics and reconstruction. Therefore, perturbing lifting features can be viewed as injecting uncertainty into {initial lifted feature fields} while keeping the learned evolution operator fixed. This converts epistemic uncertainty over weights into uncertainty over {initial conditions in feature space}, which are then propagated by a deterministic, trained operator.

\paragraph{Formalization.}
Consider a trained neural operator
\[
\mathcal{G}[\cdot;\theta]
=
\mathcal{Q}_{\theta_{\mathcal{Q}}}
\circ
\mathcal{M}_{\theta_{\mathcal{M}}}
\circ
\mathcal{P}_{\theta_{\mathcal{P}}},
\]
with optimized parameters $\hat{\theta}$.
Given an input $a$, the prediction $\hat{u}(x)$ may deviate from the (deterministic) PDE solution $u(x)$ due to epistemic uncertainty, measured by the residual field $r(x)$ in Eq.~\eqref{Equation_2}.
We define the deterministic remainder operator
\begin{equation}
\mathcal{T}
\;:=\;
\mathcal{Q}_{\hat{\theta}_{\mathcal{Q}}}
\circ
\mathcal{M}_{\hat{\theta}_{\mathcal{M}}},
\end{equation}
and treat it as a fixed surrogate of the target PDE solver after training. Epistemic uncertainty is represented by randomness injected into the lifting module. Concretely, we map the random variable(s) ${z}$ to perturbed lifting parameters $\theta_{\mathcal{P}}({z})$, and perform Monte Carlo integration in this structured subspace:
\begin{align}
\hat{u}(x)
&\approx
\frac{1}{T}\sum_{t=1}^{T}
\mathcal{T}\Big[\mathcal{P}_{\theta_{\mathcal{P}}(\mathbf{z}^{(t)})}[a]\Big](x),
\\
\mathrm{band}(x)
&\approx
\left(
\frac{1}{T}\sum_{t=1}^{T}
\Big(\mathcal{T}\big[\mathcal{P}_{\theta_{\mathcal{P}}(\mathbf{z}^{(t)})}[a]\big](x)\Big)^2
-
\hat{u}^2(x)
\right)^{\!\frac{1}{2}},
\end{align}
where $\{\mathbf{z}^{(t)}\}_{t=1}^T$ are i.i.d.\ samples and $\mathrm{band}(x)$ denotes the predictive standard deviation (a scalar half-width up to a calibrated multiplier).

\subsection{Sampling Strategies}

We propose two practical perturbation strategies that generate stochastic lifted feature fields around the deterministic features, both constructed to be zero-mean around the deterministic embedding.

\paragraph{A. Channel-wise multiplicative noise on lifted features.}\label{Method_A}
Let the lifted feature field be $\mathbf{V}_0 := \mathcal{P}[\mathbf{A}; \theta_{\mathcal{P}}]$, with channel dimension $d_v$.
For dropout rate $p\in(0,1)$, sample a {channel-wise} mask
\begin{equation}
{z}=(z_1,\dots,z_{d_v}),
\qquad
z_k \stackrel{iid}{\sim}\mathrm{Bernoulli}(1-p),
\end{equation}
and use the inverted-dropout multiplier $\xi_k := z_k/(1-p)$, so that $\mathbb{E}[\xi_k]=1$ and $\mathrm{Var}(\xi_k)=\frac{p}{1-p}$.
Broadcasting ${\xi}$ over spatial tokens yields a stochastic feature field
\begin{equation}
{V}_0^{(\mathrm{drop})}
=
\mathbf{V}_0\odot {\xi},
\end{equation}
i.e., each feature channel is multiplied by a random scalar shared across space. This makes dropout an explicit {multiplicative} perturbation in lifted feature space. Equivalently, it can be written as an {additive} zero-mean perturbation around the deterministic feature:
\begin{equation}
{V}_0^{(\mathrm{drop})}
=
\mathbf{V}_0 + {E},
\qquad
{E}:=\mathbf{V}_0\odot({\xi}-\mathbf{1}),
\qquad
\mathbb{E}[{E}]=\mathbf{0}.
\end{equation}
Entrywise, $\mathrm{Var}(E_{i,j}) = V_{0,i,j}^2\cdot \frac{p}{1-p}$, i.e., the induced perturbation is heteroscedastic and proportional to the feature magnitude.

\paragraph{Remark (noise-perturbation view as a lifting-weight subspace).}
If the lifting contains an affine map $\mathbf{V}_0=\Phi(\mathbf{A})\,\mathbf{W}_0$ (absorbing bias for simplicity), channel-wise feature dropout is algebraically equivalent to perturbing the {columns} of the lifting weights:
\begin{equation}
\mathbf{V}_0^{(\mathrm{drop})}
=
\Phi(\mathbf{A})\,(\mathbf{W}_0\mathrm{Diag}({\xi})).
\end{equation}
Thus, our procedure can be interpreted either as sampling a structured subspace of lifting weights or as sampling stochastic initial feature fields, followed by deterministic propagation through $\mathcal{T}$.

\paragraph{B. Gaussian feature perturbation.}
A simple continuous alternative is to perturb lifted features with Gaussian noise:
\begin{equation}
{V}_0^{(\mathrm{gauss})}
=
\mathbf{V}_0 + \mathbf{V}_0\odot {\epsilon}.
\end{equation}
If $\epsilon_k \stackrel{iid}{\sim}\mathcal{N}\!\left(0,\frac{p}{1-p}\right)$ (applied channel-wise and broadcast over space), then $\mathbb{E}[{V}_0^{(\mathrm{gauss})}]=\mathbf{V}_0$ while providing smooth perturbations with variance matched to the inverted-dropout scale.

\begin{figure*}[ht]
  \centering
  \includegraphics[width=\textwidth]{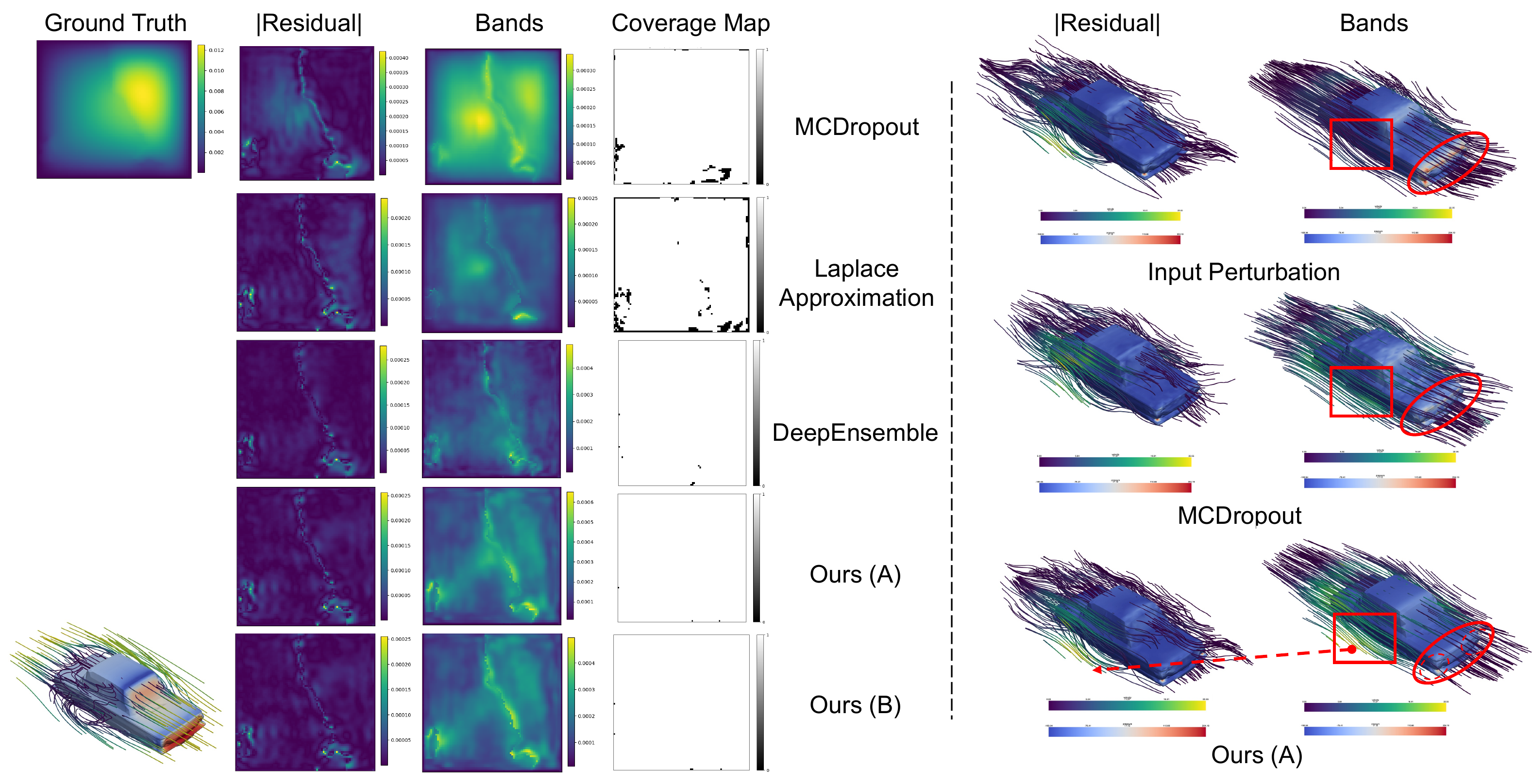}
  \caption{Visualizations of the residual and uncertainty band fields. Compared with other methods, our approach provides a more accurate characterization of the residual fields with reliable coverage.}
  \label{fig:compare_show_darcy_Car}
\end{figure*}

\begin{figure*}[ht]
  \centering
  \includegraphics[width=\textwidth]{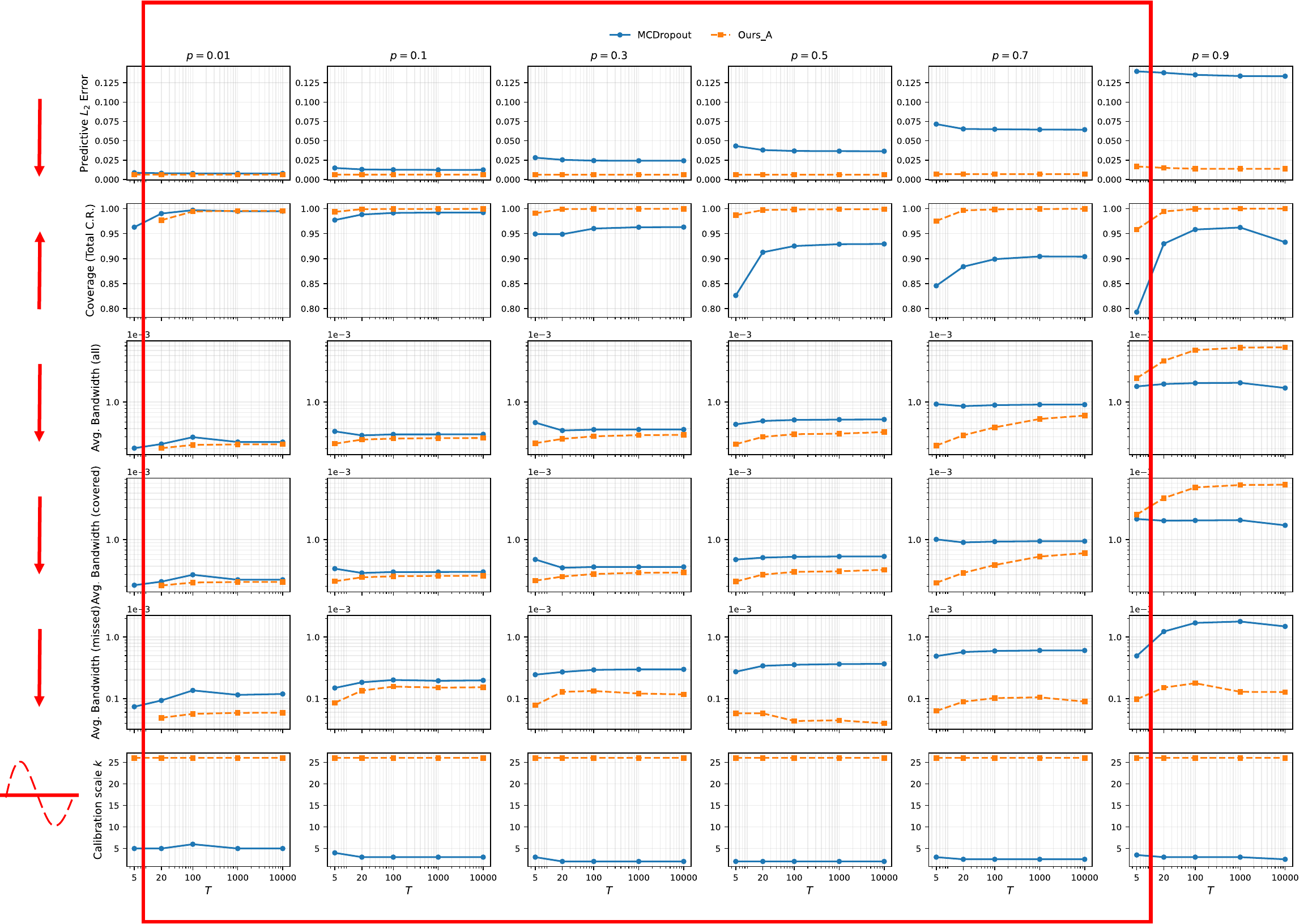}
  \caption{Comprehensive comparisons with dropout-based baselines. Notably, for each choice of $p$ and $T$, MC Dropout must be recalibrated since the uncertainty bands are altered by the newly introduced stochasticity, whereas our method~A can share a single constant calibration scale across all $p$ and $T$ settings. Note that, to obtain the case-wise Avg.\ B.W., one should multiply the reported values by the grid size $N$, which typically ranges from a few hundred to tens of thousands.}
  \label{fig:stats_MC_ours}
\end{figure*}

\begin{sidewaystable*}[ht]
\centering
\small
\setlength{\tabcolsep}{4.0pt}
\renewcommand{\arraystretch}{1.15}
\caption{Uncertainty quantification (UQ) results on two challenging benchmarks.}
\label{tab:uq_results}

\begin{tabular}{lccccc}
\toprule
\multicolumn{6}{c}{\textbf{(A) 2D Darcy Flow with Discontinuity}}\\
\addlinespace[0.35ex]
\textbf{Metric} & \textbf{MCDropout ($T=20$)} & \textbf{Laplace Approx.} & \textbf{DeepEnsemble} & \textbf{Ours (A) ($T=20$)} & \textbf{Ours (B) ($T=20$)} \\
\midrule
Total C. R. (Channel-wise C. R.) & 0.990246 & 0.869001 & \underline{0.998399} & \textbf{0.999031} & \underline{0.996810} \\
Normalized Avg. B.W. (all)                  & 2.34 & 0.77 & {1.73} & 2.73 & 2.03 \\
Normalized Avg. B.W. (covered)              & {2.35} & 0.82 & {1.73} & {2.73} & {2.03} \\
Normalized Avg. B.W. (missed)               & 0.93 & 0.48 & 1.55 & 1.35 & 1.26 \\
Time Cost of UQ &
\shortstack{$\approx$ 10 min train\\+ 1 min infer} &
\shortstack{$\approx$ 10 min train\\+ 1 min Hess.} &
\shortstack{$>$ 100 min train} &
\shortstack{$\approx$ 10 min train\\+ 0.5 min infer} &
\shortstack{$\approx$ 10 min train\\+ 0.5 min infer} \\
\bottomrule
\end{tabular}

\vspace{1.0ex}

\begin{tabular}{lccc}
\toprule
\multicolumn{4}{c}{\textbf{(B) 3D ShapeNet Car with Various Geometries}}\\
\addlinespace[0.35ex]
\textbf{Metric} & \textbf{Input Perturbation} & \textbf{MCDropout} & \textbf{Ours (A) ($T=50$)} \\
\midrule
Total C. R. (Channel-wise C. R.) &
0.5516 / 0.5656 (P) / 0.9409 (V) &
0.7761 / 0.7871 / 0.9852 &
\textbf{0.9514} / \textbf{0.9605} / \textbf{0.9905} \\
Normalized Avg. B.W. (all) &
15.432636 / 57.799948 (P) / 1.310198 (V) &
19.053221 / 71.173860 / 1.679674 &
17.702314 / 68.240521 / 0.856245 \\
Time Cost of UQ &
\shortstack{$\approx$ 6 h train\\+ 4 min infer} &
\shortstack{$\approx$ 6 h train\\+ 7 min infer} &
\shortstack{$\approx$ 6 h train\\+ 7 min infer} \\
\bottomrule
\end{tabular}

\vspace{-1.0ex}
\end{sidewaystable*}

\section{Experiments}

\paragraph{Data.} We select two challenging datasets: 2D Darcy Flow \cite{21:fno} with a discontinuous coefficient field and 3D Shape-Net Car \cite{18:shape-net-car} with various car geometries. Since Neural Operators show more pronounced prediction errors/residual fields on discontinuous fields, we use 2D Darcy Flow to evaluate how different methods estimate the complex error distributions near discontinuities. Additionally, by altering the geometric features of 3D Shape-Net Car inputs, we assess the stability of epistemic uncertainty quantification (UQ) under geometry-shifted OOD inputs.

\paragraph{Models.} We choose two representative neural operator networks: Fourier Neural Network (FNO) \cite{21:fno} and Transolver \cite{24:transolver}, both of which possess the generic architecture described in Section~\ref{Section_2_1}. In the propagation layer, FNO transfers the signal into the spectral domain for propagation, with a residual MLP layer as a bias term, while Transolver uses multi-head attention and feed-forward layers to iteratively update tensor feature fields. These architectures represent common designs in current networks, with serial (FNO) and parallel (Transolver) propagation. Our UQ methods can run directly on these architectures and are adapted to the common structure and characteristics of neural operator approximations. Furthermore, the minimal parameter budget of their lifting layers is detailed in Appendix~\ref{Appendix_parameter}. All experiments are conducted on a single \textit{Nvidia GeForce RTX 4090} GPU.

\begin{figure}[ht]
  \centering
  \includegraphics[width=0.8\textwidth]{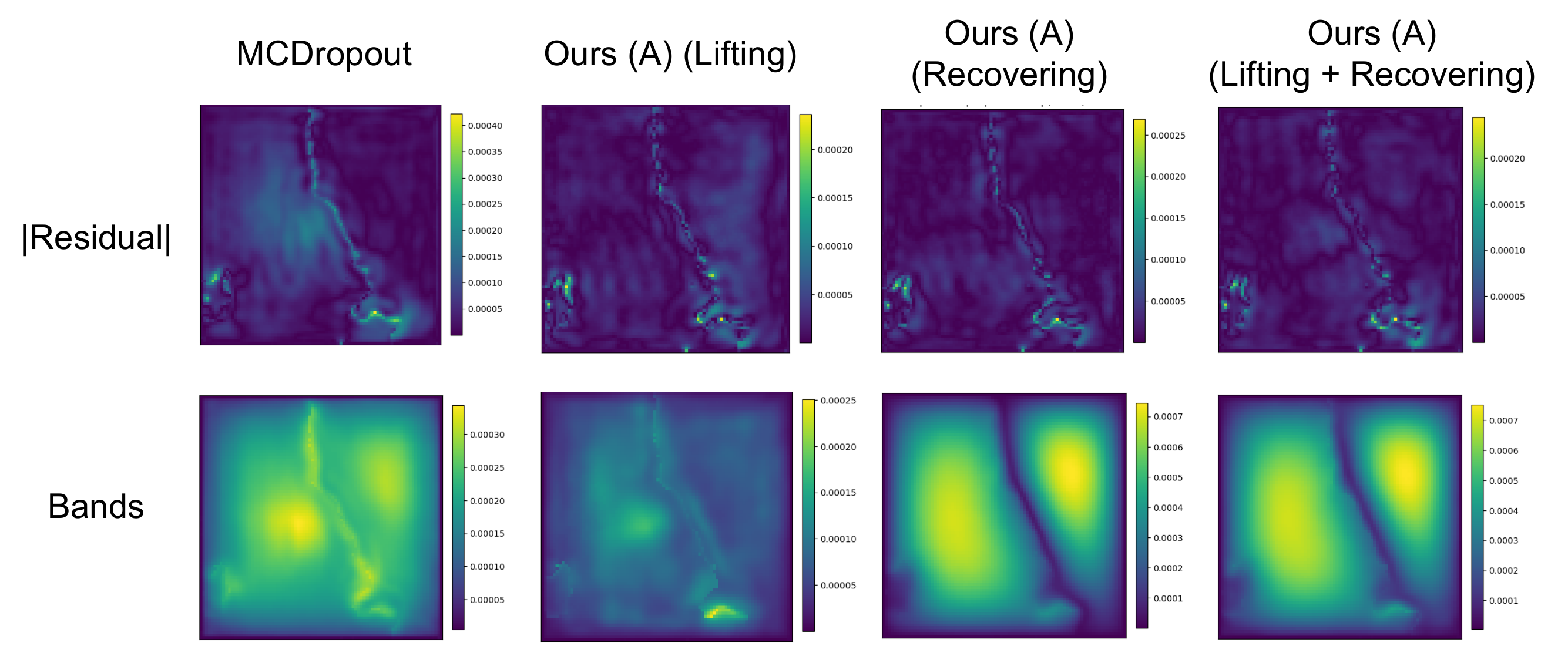}
  \caption{Demonstrations of sampling on the Lifting parameters v.s. Sampling on/with other additional choices}
  \label{fig:ablation_lift}
\end{figure}

\paragraph{Quantification Results.} Using the metrics from Section~\ref{Section_2_2} and representative visualizations, we demonstrate the advanced performance and characteristics of our method in epistemic UQ for neural operators. Specifically, we first present statistics on the entire test set, including coverage rate (Equations~\ref{eq:coverage_pointwise} and \ref{eq:coverage_global}) and bandwidth (Equation~\ref{eq:avg_bandwidth}), which reflect the performance of UQ methods. As shown in Table~\ref{tab:uq_methods_compare}, our method achieves a significantly better compression of bandwidth while maintaining a high coverage rate, compared to the stable-performing DeepEnsemble \cite{17:DeepEnsemble}, which results in much faster UQ, providing strong support for rapid design and development.

Additionally, we visualize the calibrated bands to show the actual effects of various UQ methods. The left side of Figure~\ref{fig:compare_show_darcy_Car} presents results on 2D Darcy Flow: MCDropout generates biased, conservative bands due to the loss of critical neurons, which leads to overestimation. Laplace Approximation introduces unnecessary bands in non-relevant regions, suggesting that sampling in feature spaces beyond residuals results in overestimation. DeepEnsemble is more robust than these two methods in capturing the spatial distribution of residuals. Our method, requiring significantly less time (see Table~\ref{tab:uq_methods_compare}), achieves comparable even better coverage than DeepEnsemble.

The right side shows uncertainty quantification results for the 3D car wind-tunnel simulation (Appendix~\ref{bm:shapenetcar}). Input noise injection \cite{24:uncertainty_workshop_NO} on the original input fails to capture the residual structure in the vector velocity field and yields inflated uncertainty on the pressure field. MCDropout is affected by biased mean predictions and redundant stochastic hypotheses; because dropout perturbs all neurons symmetrically, many samples become weakly informative and tend to average out localized turbulent patterns around the car body, resulting in globally widened and spatially smeared bands. In contrast, our method A injects stochasticity only at the lifted feature level by perturbing the learned input embeddings while keeping the trained operator fixed. This produces bands that better align with the spatial organization of residuals, with improved coverage in regions such as the headlamp pressure residuals and the side-flow velocity trends. Overall, our method provides designers with more reliable localization of epistemic uncertainty and appropriately conservative uncertainty estimates.

\paragraph{Comprehensive comparison with dropout-based baselines.}
We comprehensively compare our method~A against MCDropout in Figure~\ref{fig:stats_MC_ours}. Across a wide spectrum of dropout probabilities $p$ and numbers of forward runs $T$, our method~A achieves better prediction accuracy, higher (or comparable) coverage, lower bandwidth, and substantially reduced sensitivity to the calibration scale. The main exceptions occur in degenerate regimes, namely very large $p>0.9$ or very small $T\le 5$, where effective stochasticity can collapse (i.e., multiple stochastic forward passes become nearly identical), as illustrated in Figure~\ref{fig:Ours_A_T_p_01}. Overall, these results suggest that applying dropout indiscriminately to neural operators can introduce biased predictions and miscalibrated uncertainty (often manifesting as overly conservative or spatially smeared bands), whereas our structure-aware dropout sampling yields a more stable and reliable UQ mechanism for scientific computing.

\paragraph{Ablations.} We perform an ablation study by changing the sampling module, based on our method A implementation, and applying dropout at different subspaces in the parameter space across different layers. As shown in Figure~\ref{fig:ablation_lift}, sampling exclusively in the lifting subspace ensures faithful uncertainty quantification.

\section{Conclusion}
We presented a structure-aware approach to epistemic uncertainty quantification for neural-operator PDE surrogates, motivated by the observation that generic, structure-agnostic perturbations can be statistically inefficient and can distort residual structures that UQ is meant to capture. By sampling uncertainty {only} through the lifting module, interpretable as perturbing the initial lifted feature fields while keeping the learned evolution operator fixed, our method produces uncertainty bands that better track localized residual patterns, improving reliability for downstream scientific-computing workflows. Empirically, across representative benchmarks and geometrically OOD stress tests, this design improves calibration and spatial interpretability relative to standard MCDropout, Laplace-style approximations, and input perturbation, while remaining lightweight at inference time.

This work is primarily empirical and deployment-oriented rather than theoretical: our goal is a practical UQ mechanism that can be integrated into existing neural operators with minimal modification. Promising future directions include exploring richer choices of the sampling variable $z$ (beyond the current feature-noise instantiations) and extending the structure-aware principle to coupled multi-physics and multi-field operators, where uncertainty must be propagated consistently across interacting state variables.

\section{GenAI Disclosure}

This manuscript was prepared with the assistance of a large language model (LLM) to improve the clarity, grammar, and overall fluency of sentences and paragraphs. The LLM was used solely for language polishing and stylistic refinement and did not contribute to the scientific content, technical ideas, data analysis, experimental design, or conclusions. All results, interpretations, and claims presented in this work are entirely those of the authors, who take full responsibility for the content.

\newpage
\appendix
\section{Neural Operators}

\paragraph{Loss} \label{Appendix_loss} We aim to approximate the operator $\mathcal{G}$ in Section \ref{Section_2_1} by optimizing the model parameters $\theta \in \Theta$ through relative $L_2$ loss below:
\begin{align} \label{loss_l_2}
    \min_{\theta \in \Theta} \mathcal{L}(\theta) &:= \min_{\theta \in \Theta} \frac{1}{M} \sum_{m=1}^{M} \left[ \frac{\lVert {\mathcal{G}}_{\theta}[a_m] - u_m\rVert_2}{\lVert u_m \rVert_2}  \right ] \\
    & = \min_{\theta \in \Theta} \frac{1}{M} \sum_{m=1}^{M} \left[ \frac{\lVert {\mathcal{G}}[\mathbf{A}_m; \theta] - \mathbf{U}_m\rVert_F}{\lVert \mathbf{U}_m \rVert_F}  \right ],
\end{align}
where $\Theta$, $M$, and $u_m=u_m(x)$ denote the parameter space, the number of function samples (in one batch), and the $m$-th output function in $M$ samples, respectively.

\paragraph{Parameter allocation}\label{Appendix_parameter}
Since the lifting operator $\mathcal{P}[\cdot;\theta_{\mathcal{P}}]$ is typically implemented as a channel-wise fully-connected map, its parameter budget is negligible compared with the remaining blocks. This remains true even when positional encoding (e.g., concatenated absolute or reference coordinates) is included in the input channels.

For instance, on 2D Darcy, the lifting layer of an FNO contains only $128$ parameters out of $1{,}196{,}801$ in total, accounting for just $0.0107\%$ of the model. In contrast, architectures with heavier input embeddings (e.g., Transolver) allocate a larger yet still relatively small fraction to the lifting stage: Transolver uses $16{,}896$ lifting parameters out of $2{,}826{,}945$ total, i.e., $0.598\%$.

\section{Benchmarks}

Here we select two challenging benchmarks: one with discretized fields on a uniform mesh, and another with fields defined on a non-uniform grid featuring out-of-distribution (OOD) car geometries.

\paragraph{2D Darcy Flow} \label{bm:2ddarcy} The 2D Darcy Flow is a second-order elliptic equation as follows:
\begin{align}
    -\nabla \cdot(a\nabla u) &= f,
\end{align}
where $f=1$. Our goal is to approximate the mapping operator from the spatially-distributed coefficient field $a({x})$ to the solution function $u({x})$ with zero Dirichlet boundary conditions \cite{21:fno}. Models are trained on 1000 training samples, which are down-sampled to a size of ($85 \times 85$), and tested on 100 samples.

\paragraph{3D ShapeNet Car} \label{bm:shapenetcar}
The 3D ShapeNet Car benchmark targets aerodynamic performance prediction for vehicle geometries from the ShapeNet car subset, using CFD simulations conducted at a fixed inlet speed of 72 km/h \cite{18:shape-net-car}. Given a car shape (represented on a fixed computational mesh), the task is to learn the surrogate operator that maps geometry-conditioned inputs to the corresponding 3D flow fields—specifically the velocity field and pressure field. The dataset contains 889 simulated samples, split the out of distribution 100 cases with unseen shapes for testing.
\section{Limitations and Ethical Considerations}
\paragraph{Limitations.}
(i) {Scope of uncertainty:} We focus on epistemic uncertainty in largely deterministic PDE settings; the method does not explicitly model aleatoric uncertainty arising from internal numerical errors of the reference solver (e.g., discretization and iterative tolerance effects).  
(ii) {Calibration transfer:} The calibration scale is tuned on held-out data; under severe distribution shift, calibrated bands may no longer maintain nominal coverage, so additional OOD detection, periodic recalibration, or conservative fallback rules may be required.
(iii) {Empirical focus:} This work is driven by practical deployability and empirical evidence, rather than formal bounds under reasonable assumptions. A probabilistic theoretical exploration for general neural operators is left for future work.

\paragraph{Ethical considerations.}
We do not foresee explicit ethical issues specific to this work. The proposed method improves uncertainty reporting for scientific-computing surrogates and is intended to support more reliable deployment (e.g., flagging cases for additional high-fidelity simulation), rather than replacing established verification and validation practices.

\section{Epistemic Uncertainty Quantification}

\subsection{Showcases}

\begin{figure*}[ht]
  \centering
  \includegraphics[width=\textwidth]{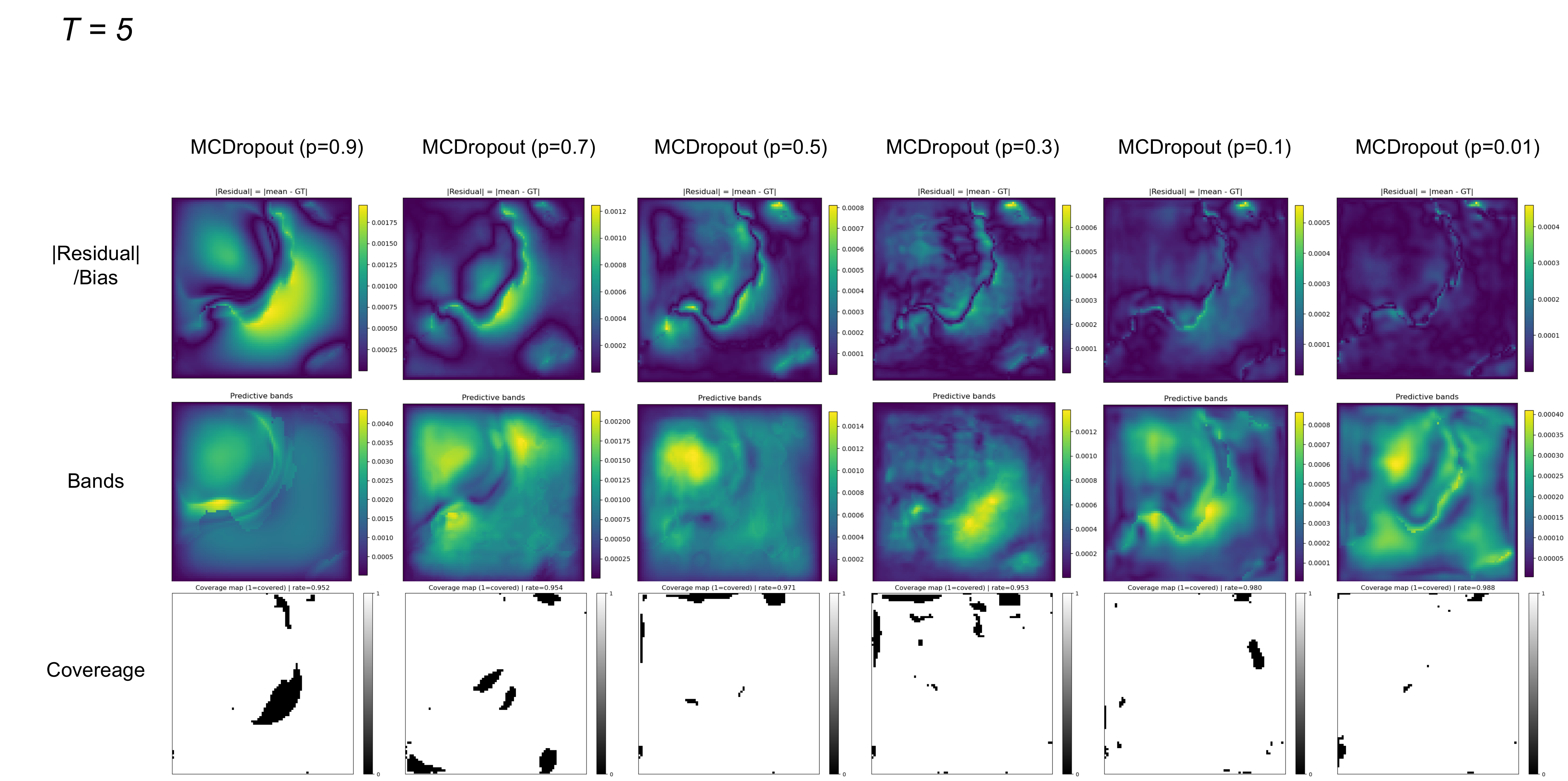}
    \includegraphics[width=\textwidth]{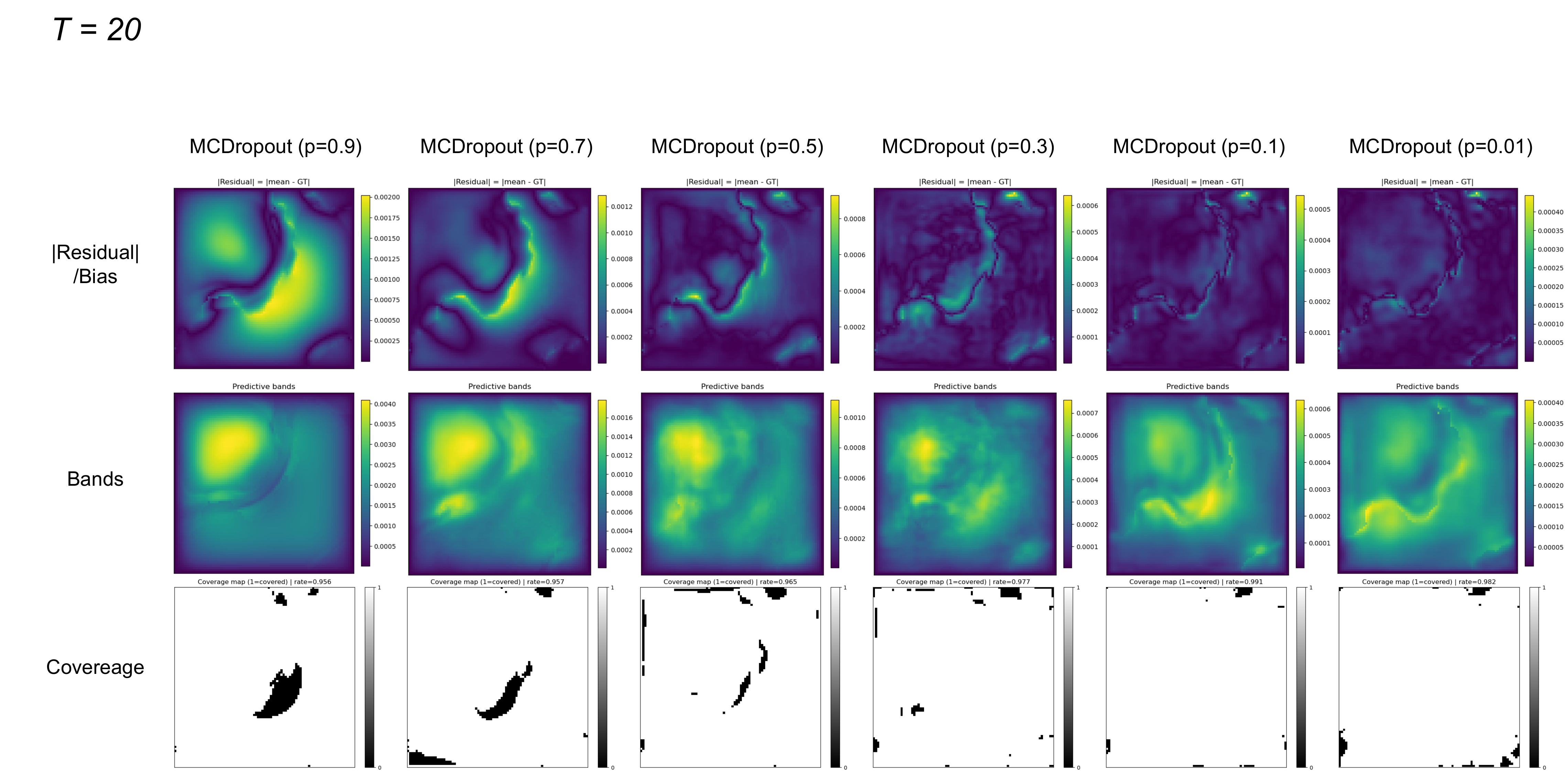}
  \caption{Naive MCDropout uncertainty for different dropout probabilities $p$ and sample counts $T$. Higher $p$ causes more biased predictions and lower coverage of the calibrated uncertainty bands.}
  \label{fig:MCDropout_T_p_01}
\end{figure*}

\begin{figure*}[ht]
  \centering
  \includegraphics[width=\textwidth]{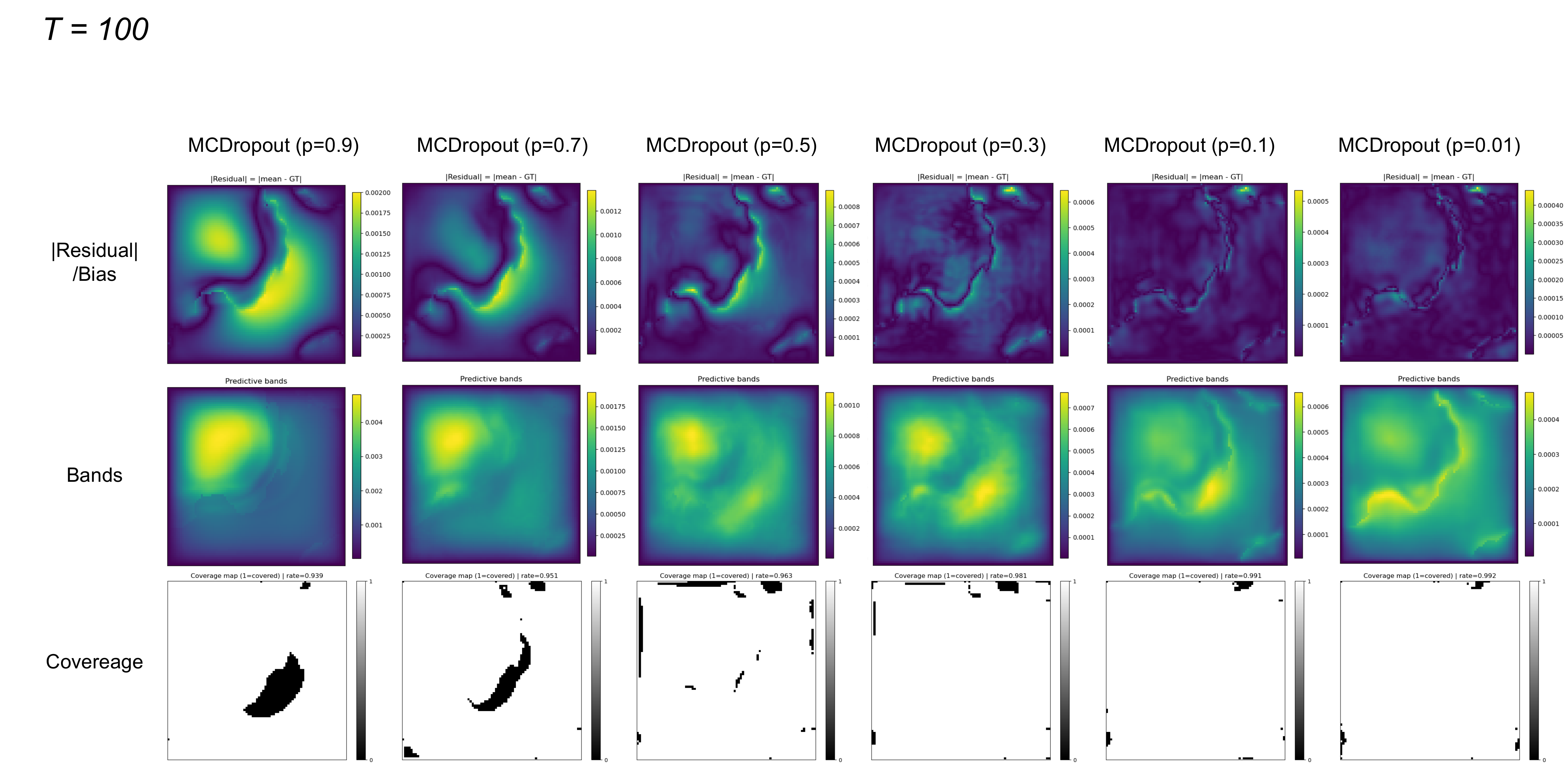}
    \includegraphics[width=\textwidth]{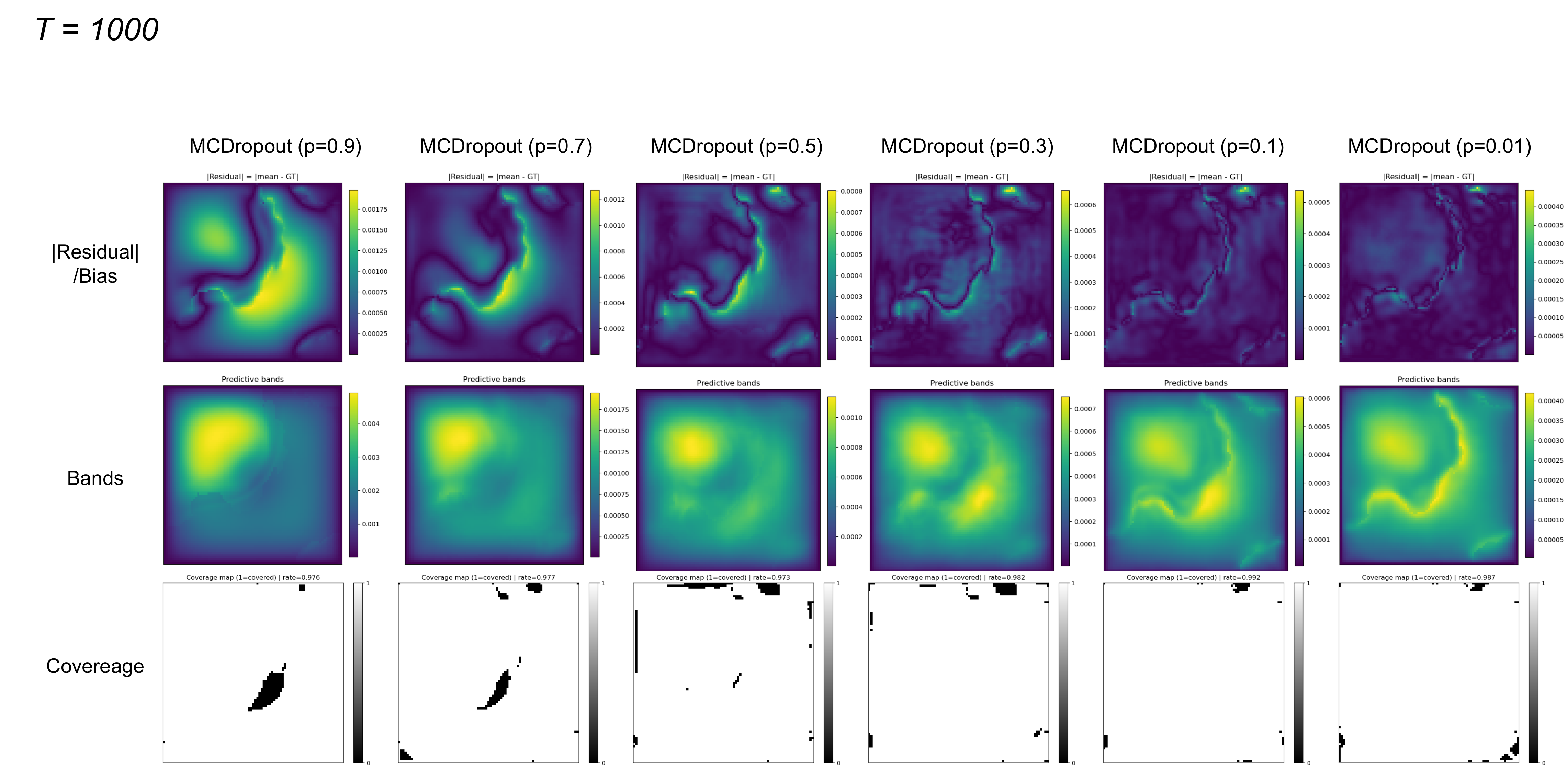}
  \caption{Naive MCDropout uncertainty for varying dropout probability $p$ and sample counts $T$. Relative to the smaller-$T$ results in Figure~\ref{fig:MCDropout_T_p_01}, increasing $T$ yields a smoother, re-distributed uncertainty-band field.}
  \label{fig:MCDropout_T_p_02}
\end{figure*}

\begin{figure*}[ht]
  \centering
  \includegraphics[width=\textwidth]{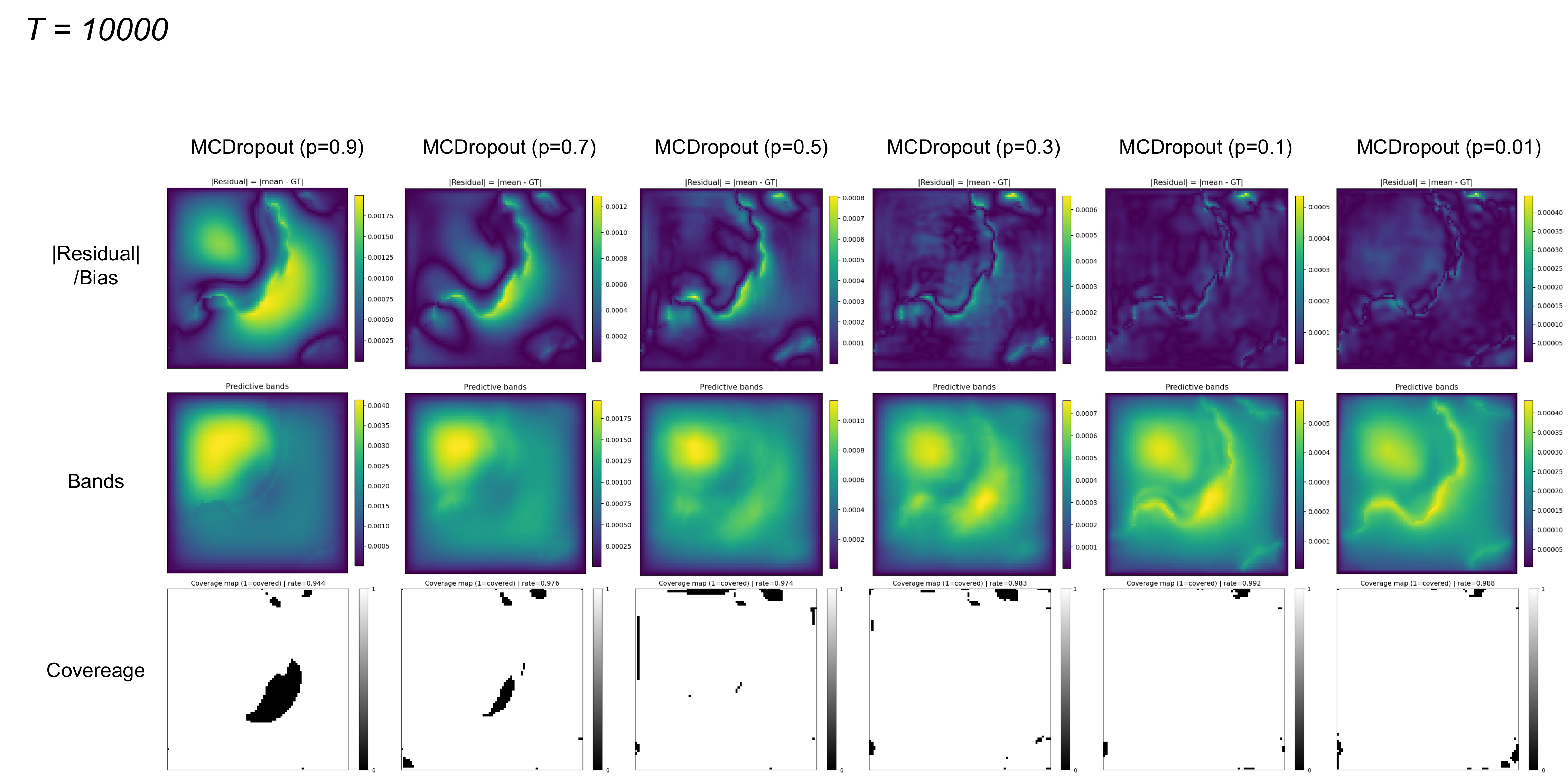}
  \caption{Naive MCDropout uncertainty for varying dropout probability $p$ with a relatively large sample count $T=10,000$. The near-identical predictive means, uncertainty bands, and coverage indicate that further increasing $T$ has a negligible impact on the estimated mean and uncertainty band.}
  \label{fig:MCDropout_T_p_03}
\end{figure*}

\begin{figure*}[ht]
  \centering
  \includegraphics[width=\textwidth]{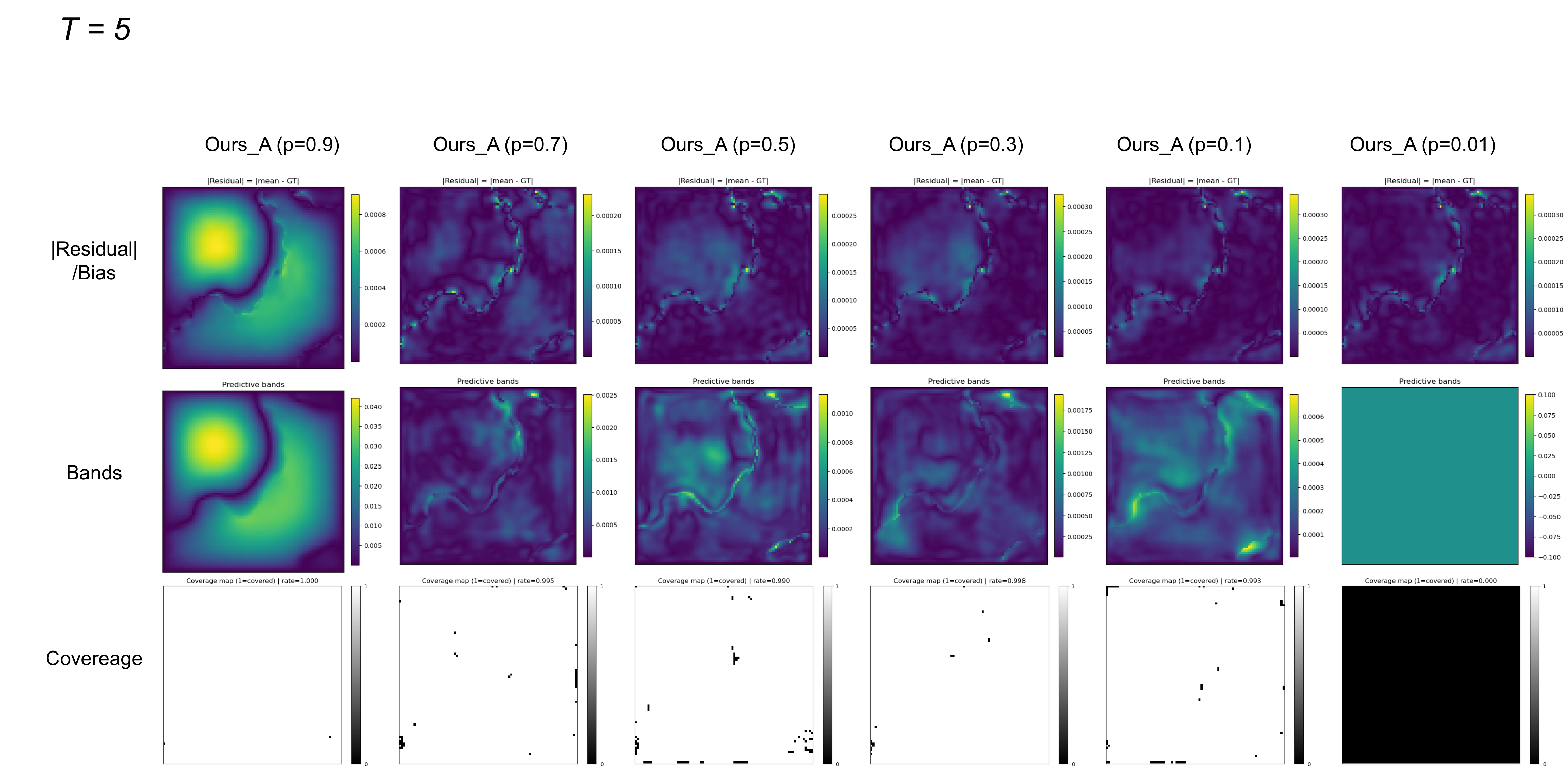}
  \includegraphics[width=\textwidth]{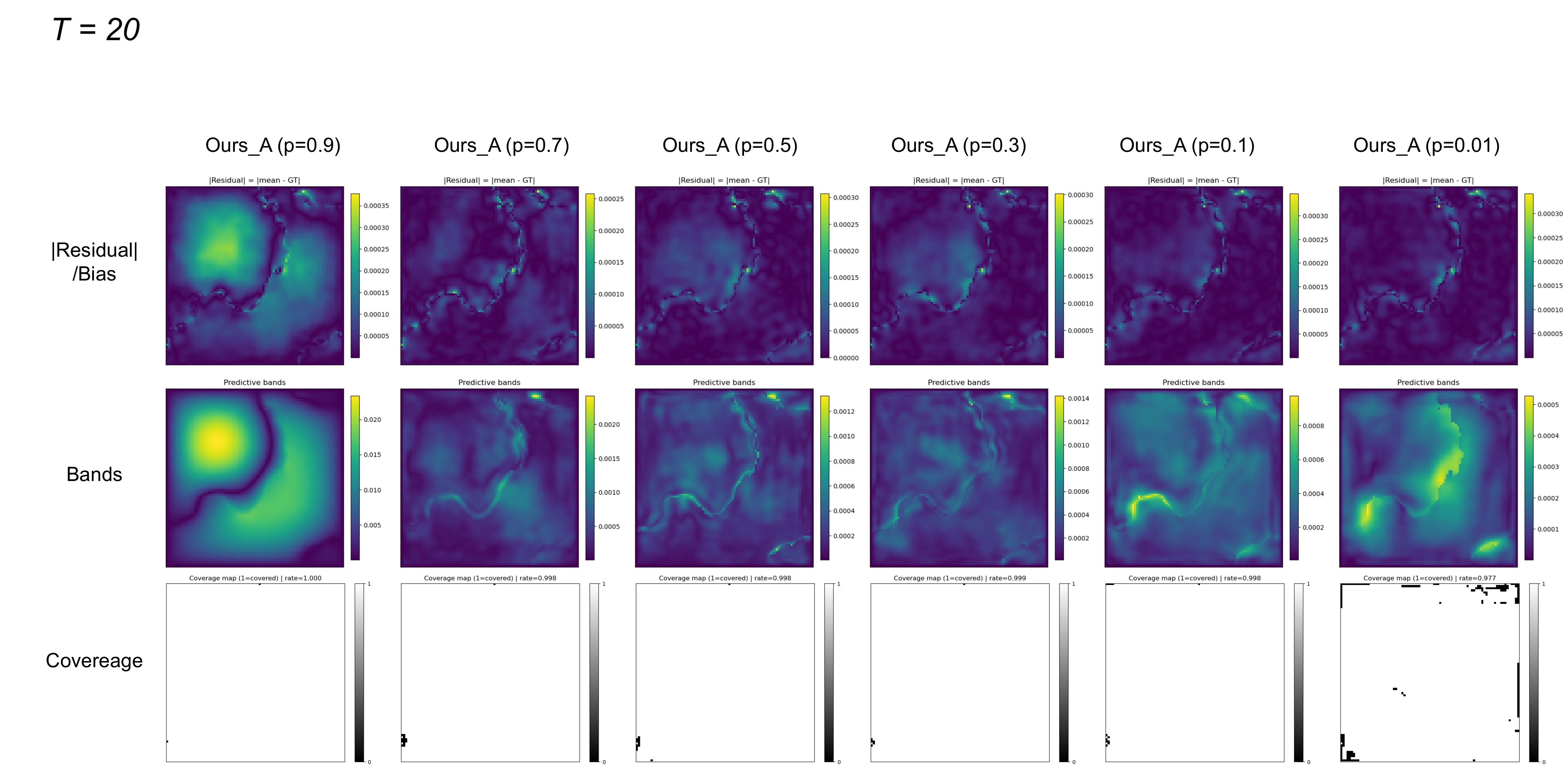}
  \caption{Quantified uncertainty visualizations of the A-implementation of our method for varying dropout probabilities $p$ and sample counts $T$. With an appropriate dropout rate (i.e., neither extremely small nor large), our approach yields consistently far less biased predictions and higher coverage with calibrated uncertainty bands than the naive MCDropout results in Figure~\ref{fig:MCDropout_T_p_01}. Notably, the calibration scale $k$ is shared across all $p$ values in our method A, whereas naive MCDropout typically requires case-by-case tuning of $k$.}
  \label{fig:Ours_A_T_p_01}
\end{figure*}

\begin{figure*}[ht]
  \centering
  \includegraphics[width=\textwidth]{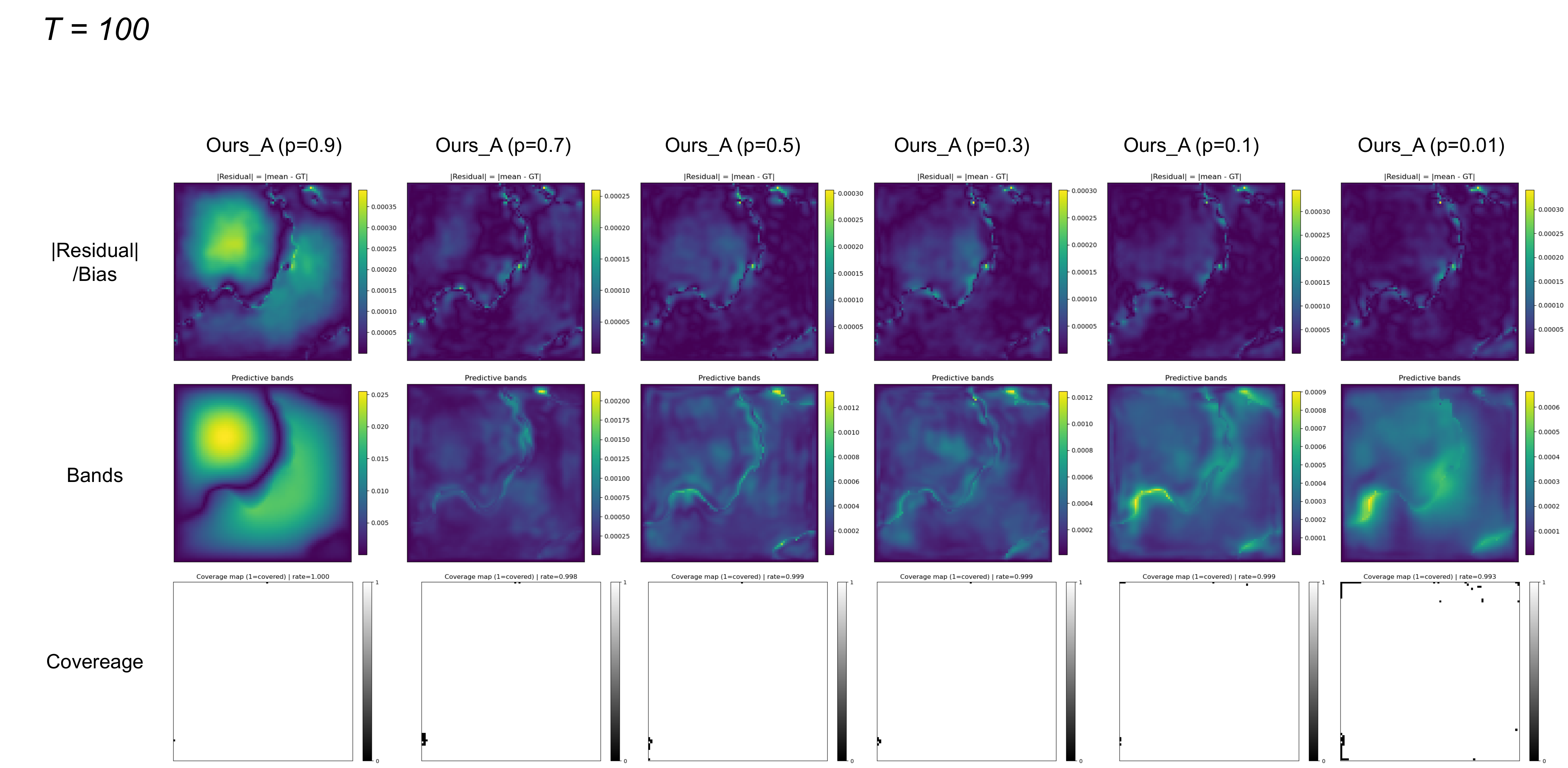}
  \includegraphics[width=\textwidth]{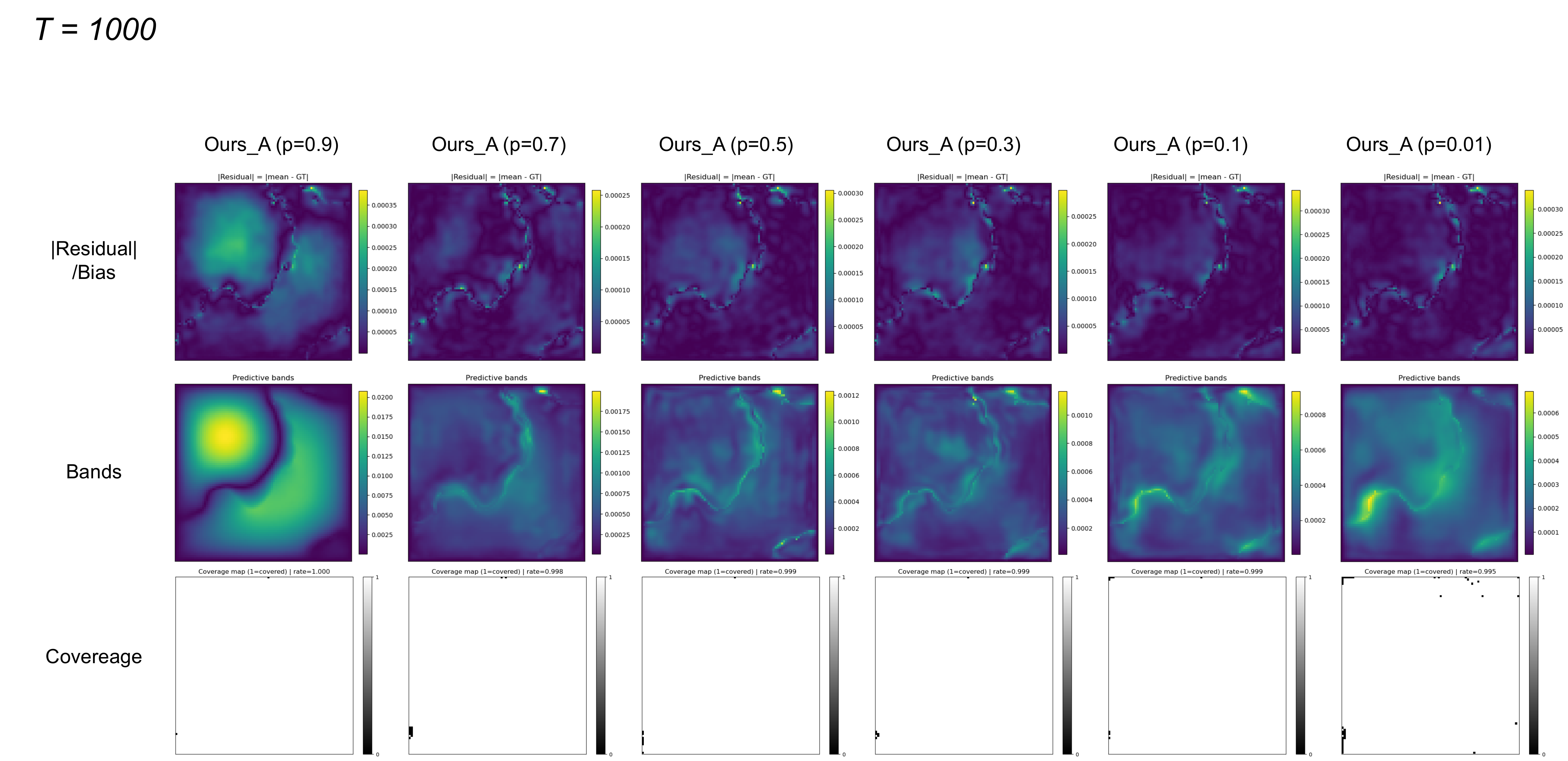}
  \caption{Quantified uncertainty visualizations of the A-implementation of our method for varying dropout probabilities $p$ and sample counts $T$. An appropriate dropout rate (i.e., neither extremely small nor large) provides reliable uncertainty quantification even with small $T$ shown in Figure~\ref{fig:Ours_A_T_p_01}. Increasing $T$ to $100$ or $1000$ offers little benefit for normal $p$ but helps reduce under-coverage when the dropout probability is very small ($p=0.01$).}
  \label{fig:Ours_A_T_p_02}
\end{figure*}

\begin{figure*}[ht]
  \centering
  \includegraphics[width=\textwidth]{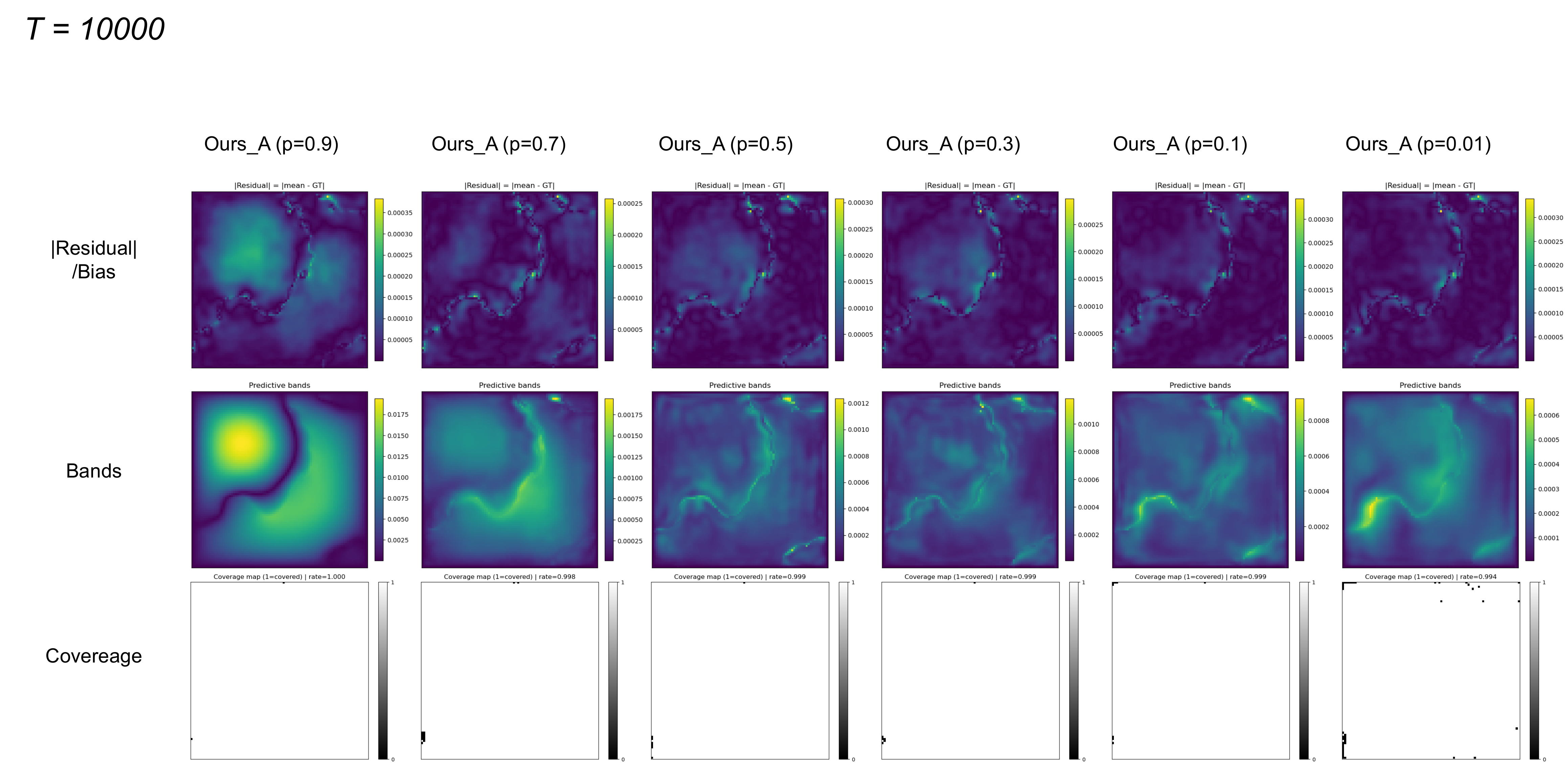}
    \caption{Quantified uncertainty visualizations of the A-implementation of our method for varying dropout probabilities $p$ and sample counts $T$. Compared with the previously small $T$ cases, further increasing $T$ to $10000$ has fewer benefits on coverage. However, compared with the uncertainty bands in Figure~\ref{fig:MCDropout_T_p_03}, the A-implementation of our method provides a more faithful estimation of the residual field and achieves consistently much higher coverage.}
  \label{fig:Ours_A_T_p_03}
\end{figure*}

\paragraph{MCDropout vs.\ Ours-A.}
MCDropout is implemented by inserting dropout layers before every linear layer / linear operator. We visualize the predictive mean and uncertainty band fields of MCDropout~\cite{16:dropout} and our implementation of Method~A (Section~\ref{Method_A}) under different numbers of samples $T$ and dropout probabilities $p$ in Figures~\ref{fig:MCDropout_T_p_01}--\ref{fig:MCDropout_T_p_03} and Figures~\ref{fig:Ours_A_T_p_01}--\ref{fig:Ours_A_T_p_03}. The corresponding coverage maps are shown alongside each set of results.

% Bibliography (switch style as you like)
\bibliographystyle{plain}
\bibliography{reference}

% Your extra figure/code inputs

\begin{figure*}[ht]
  \centering
  \includegraphics[width=\textwidth]{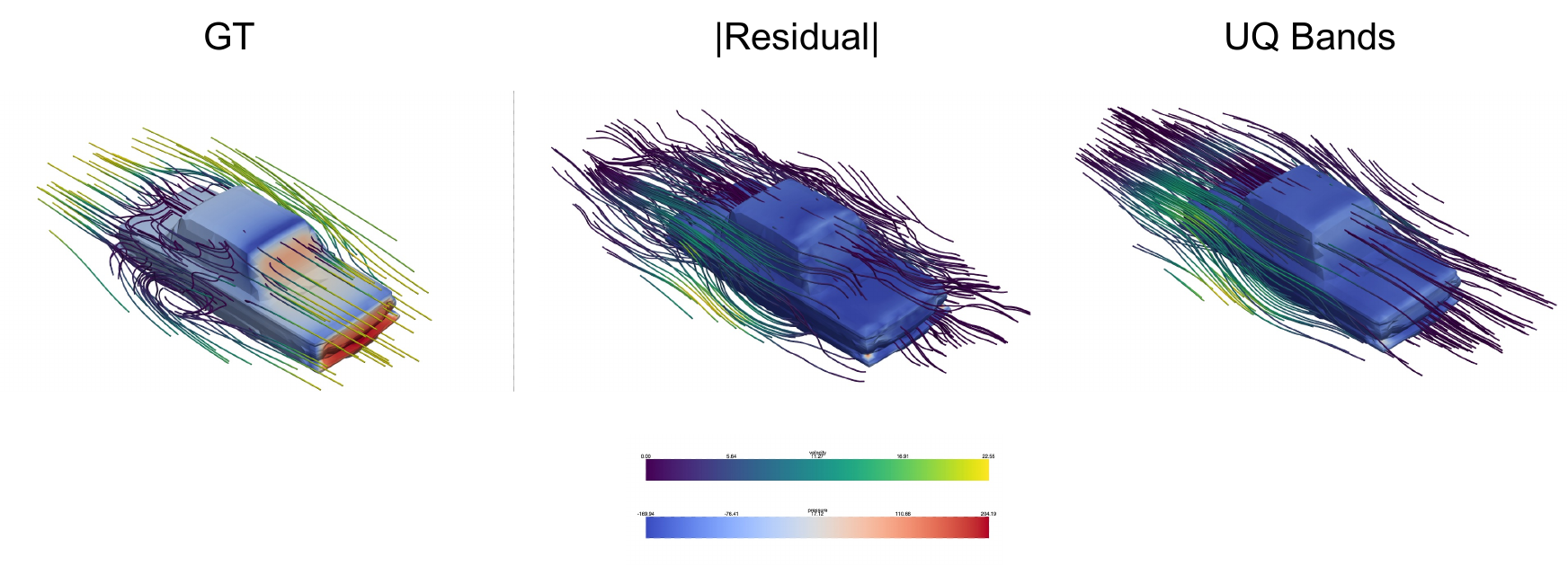}
    \caption{UQ visualizations of our method B on the \textit{3D ShapeNet Car} with C.R.: 0.9162 (\textit{All}) / 0.9248 (\textit{Press}) / 0.9896 (\textit{Velocity}) and Normalized Avg. B.W.: 17.530033 (\textit{All}) / 0.970439 (\textit{Press}) / 67.208816 (\textit{Velocity}).}
  \label{fig:Our_B_Car}
\end{figure*}

\end{document}